\documentclass[runningheads]{llncs}

\usepackage{eccv}

\usepackage{eccvabbrv}

\usepackage{graphicx}
\usepackage{booktabs}
\usepackage{times}
\usepackage{multirow}
\usepackage{epsfig}
\usepackage{amsmath}
\usepackage{amssymb}
\usepackage{booktabs}
\usepackage{verbatim}
\usepackage{wrapfig}
\usepackage{enumitem}
\usepackage{pifont}
\usepackage{algpseudocode}%
\usepackage{listings}%
\usepackage{bbding}
\usepackage[accsupp]{axessibility}  

\usepackage{hyperref}

\usepackage{orcidlink}

\begin{document}

\title{LiveHPS++: Robust and Coherent Motion Capture \\in Dynamic Free Environment}



\author{Yiming Ren\inst{1} \and
Xiao Han\inst{1} \and
Yichen Yao\inst{1} \and
Xiaoxiao Long\inst{2} \and 
Yujing Sun\inst{2\star} \and\\
Yuexin Ma\inst{1}\thanks{Corresponding author. This work was supported by NSFC (No.62206173), MoE Key Laboratory of Intelligent Perception and Human-Machine Collaboration (ShanghaiTech University), Shanghai Frontiers Science Center of Human-centered Artificial Intelligence (ShangHAI).}}

\authorrunning{Ren, Y. et al.}

\institute{$^1$ShanghaiTech University, $^2$The University of Hong Kong \email{\{renym2022,mayuexin\}@shanghaitech.edu.cn}}
\maketitle
\begin{figure}[ht!]
	\centering
	\includegraphics[width=\linewidth]{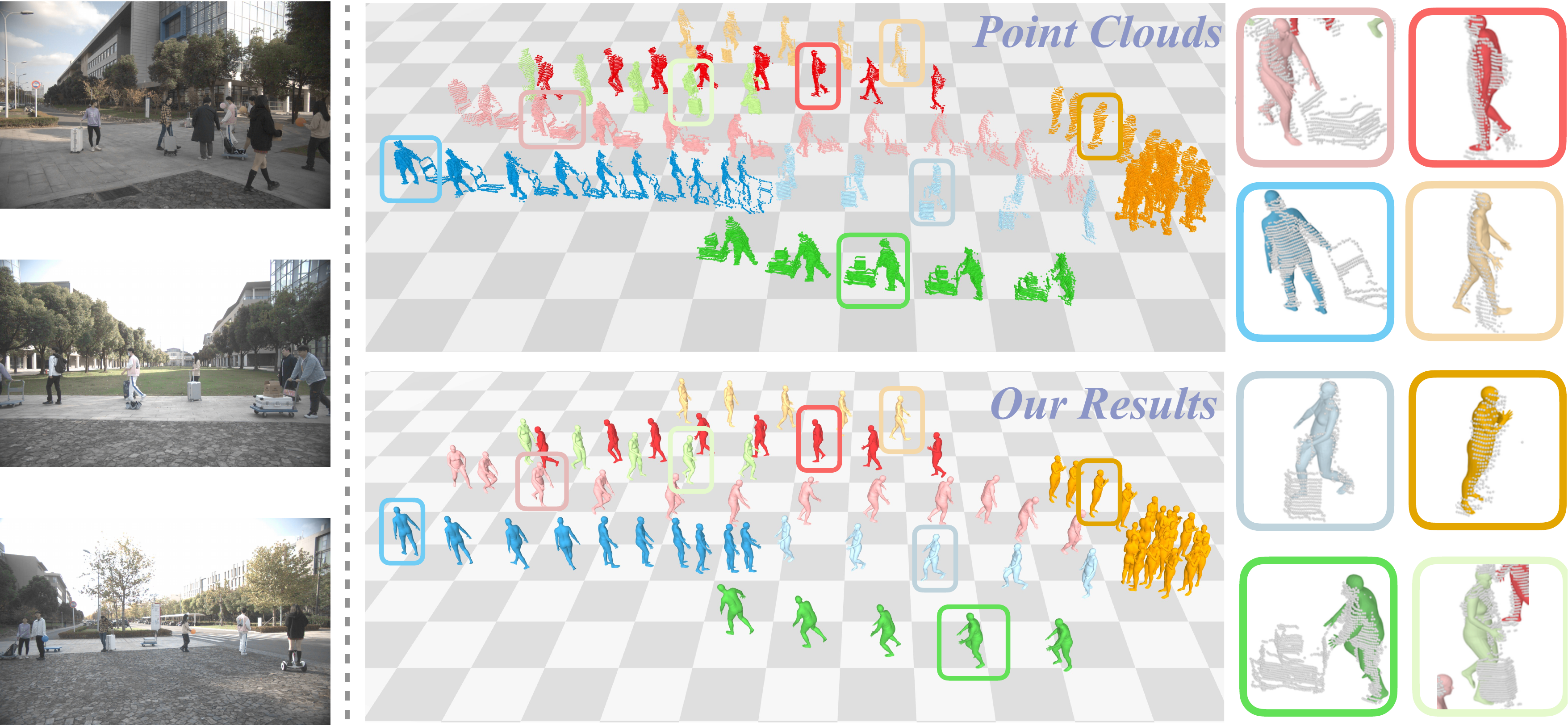}
        \vspace{-4ex}
	\caption{Visualization of the motion capture performance of LiveHPS++ in a real-time captured scenario with complex human-object interaction. The left exhibits images for reference, the middle shows the noised point clouds (top) and our corresponding mesh model results (bottom). We zoom in some cases on the right for clearer demonstration, where point clouds are drawn in white.}
	\label{fig:teaser}
	\vspace{-6ex}
\end{figure}

\begin{abstract}


LiDAR-based human motion capture has garnered significant interest in recent years for its practicability in large-scale and unconstrained environments. However, most methods rely on cleanly segmented human point clouds as input, the accuracy and smoothness of their motion results are compromised when faced with noisy data, rendering them unsuitable for practical applications.
To address these limitations and enhance the robustness and precision of motion capture with noise interference, we introduce LiveHPS++, an innovative and effective solution based on a single LiDAR system. Benefiting from three meticulously designed modules, our method can learn dynamic and kinematic features from human movements, and further enable the precise capture of coherent human motions in open settings, making it highly applicable to real-world scenarios.
Through extensive experiments, LiveHPS++ has proven to significantly surpass existing state-of-the-art methods across various datasets, establishing a new benchmark in the field. \url{https://4dvlab.github.io/project_page/LiveHPS2.html}

\end{abstract}
\section{Introduction}
\label{sec:intro}
Capturing accurate and natural human motions in large-scale dynamic environments is pivotal for the modeling and analysis of human behaviors, as well as for enhancing the understanding of 3D scenes. This foundational work significantly benefits many downstream applications, ranging from digital filmmaking and delivering immersive experiences in AR/VR gaming to training robots to emulate human-like behaviors and effectively collaborate with humans. Previous motion capture methods are usually based on optical devices~\cite{AminARS2009,BurenSC2013,ElhayAJTPABST2015,RhodiRRST2015,Robertini:2016,Pavlakos17,Simon17} or wearable devices~\cite{XSENS,noitom,yi2021transpose,PIPCVPR2022}. However, the former is sensitive to light conditions and not suitable for outdoor scenarios, and the latter requires the actor to wear a body-mounted IMU suit, not applicable for daily-life usage and fails to capture human shapes.

LiDAR has emerged as a foundational sensor in the realms of robotics and autonomous driving~\cite{zhu2021cylindrical,zhu2020ssn,cong2022stcrowd,xu2023human}, due to its exceptional long-range depth-sensing capabilities. Notably, LiDAR point clouds can offer precise 3D geometry and location information of the human body without limitation of light conditions or wearable devices. This makes LiDAR very promising for tracking how people move and act in free environment.
Recently. some advancements~\cite{li2022lidarcap,jang2023movin,cai2023pointhps} have already underscored the efficacy of single-LiDAR systems for capturing human motion. However, these techniques are primarily effective with clean human point clouds within controlled experimental settings, and they often fall short when confronted with the intricacies of real-world application scenarios with noise and occlusions. 


To address these problems, LiveHPS~\cite{ren2024livehps} has collected a vast dataset of human motion captured in real-world settings, featuring natural interactions with other people, and has also put forth an effective strategy to solve the varying point distributions stemming from occlusions and noise. Nevertheless, it treats features from real human points and noise points equally and ignores the coherence of global poses and translations, causing the accuracy and consistency of its predicted motions to face limitations in complex scenarios, where the noise does not just come from the sensor accuracy but also arises from the horrible segmentation results by upstream perception algorithms. As illustrated in Fig.\ref{fig:teaser}, distinguishing clean human point clouds becomes particularly challenging when individuals are in close proximity or interacting with objects. This usually results in disastrous inputs with much noise for motion capture algorithms, leading to inaccurate and jerky motion outcomes. Enhancing the robustness and precision of LiDAR-based motion capture methods in any complicated situation becomes very crucial for real-world deployment and applications.

In this paper, we introduce \textbf{LiveHPS++}, an innovative and effective approach for capturing precise and coherent human motions across vast, unregulated environments using a single LiDAR system. Our method consists of three specially designed modules to tackle above challenges. Beginning with sequential point cloud inputs, the first module, the \textbf{Trajectory-guided Body Tracker}, subtly captures the dynamic characteristics of human movement through their trajectories, ensuring consistency across adjacent poses. Following this, the second module, the \textbf{Noise-insensitive Velocity Predictor}, employs a cross-attention mechanism to make each human joint engage with the most relevant point features, minimizing the impact of extraneous noise. This module forecasts the velocity of each joint, explicitly modeling the kinematic details of the human motion, which is very valuable for refining the global pose and translation in the third module, the \textbf{Kinematic-aware Pose Optimizer}. Here, a sophisticated system of candidate generation and feature interaction is implemented to achieve more precise pose adjustments. In particular, we propose a synthesized motion dataset, \textbf{NoiseMotion}, built on SURREAL~\cite{varol2017learning} and ShapeNet~\cite{chang2015shapenet} to enlarge the challenging noised data in complex scenarios by simulating human interactions with various objects. Benefiting from both the new dataset and our effective algorithm, we significantly enhance the robustness and accuracy of motion capture in complex settings, where noise interference is prevalent, thereby facilitating the practical deployment of this technology. 

Extensive experiments are conducted on multiple LiDAR-based motion datasets, including NoiseMotion, FreeMotion~\cite{ren2024livehps}, FreeMotion-OBJ~\cite{ren2024livehps}, and Sloper4D~\cite{dai2023sloper4d}. Compared with the state-of-the-art method, LiveHPS++ has improved the performance by a large margin, e.g. 6.28\% and 69.29\% for the global vertex error and the jitter on FreeMotion-OBJ, the most challenging real human motion dataset, and 23.05\% and 13.54\% for the global vertex error and the jitter on NoiseMotion, the most challenging synthetic human motion dataset. Moreover, detailed ablation studies and discussions are also provided to verify the effectiveness of detailed designs in our network. Contributions of this work can be summarized as follows.

\begin{itemize}
    \item[$\bullet$] We propose a robust human motion capture method, LiveHPS++, which can eliminate the effect of severe noises and produce precise and natural human motions in dynamic free environment, which is very practical for real-world applications.
    \item[$\bullet$] We design three effective modules in our method, which can implicitly and explicitly model dynamic and kinematic features of human motions to facilitate the coherence and accuracy of motion capture results.
    \item[$\bullet$] LiveHPS++ achieves state-of-the-art performance on various datasets and significantly outperforms existing methods.
\end{itemize}

\section{Related Work}
\label{sec:related_work}
\subsection{Optical-based Motion Capture}
Early systems of optical-based motion capture, characterized by their reliance on camera-tracked markers to reconstruct 3D meshes~\cite{VICON,Vlasic2007,optitrack}, laid the groundwork for high-quality motion capture, becoming a staple in professional settings. 
The field has seen substantial advancements with the introduction of markerless mocap technologies~\cite{BreglM1998,de2008performance,TheobASST2010,FlyCap,HolteTTM2012,SigalBB2010,SigalIHB2012,StollHGST2011,JooLTGNMKNS2015,UnstructureLan,yuan2020residual,luo2021dynamics}. These innovations offer a less intrusive and often more cost-effective solution, leveraging multi-view algorithms to maintain robustness even in uncontrolled environments~\cite{AminARS2009,BurenSC2013,ElhayAJTPABST2015,RhodiRRST2015,Robertini:2016,Pavlakos17,Simon17}. 
However, the complexity of synchronizing and calibrating multi-camera setups remains a challenge. In response, monocular mocap methods have been developed, employing a range of techniques from optimization and regression~\cite{TAM_3DV2017,Lassner17,bogo2016keep,Kolotouros_2019_CVPR,HMR18,Kanazawa_2019CVPR,VIBE_CVPR2020,zanfir2020neural} to template-based and probabilistic approaches~\cite{MonoPerfCap,LiveCap2019tog,EventCap_CVPR2020,DeepCap_CVPR2020,challencap}. 
To address depth ambiguity, some researchers have turned to depth cameras~\cite{Shotton:2011,Baak:2011,Wei:2012,DoubleFusion,guoTwinFusion}, which, despite their benefits, are limited by their sensing range and perform poorly in outdoor settings. 
Nonetheless, optical-based motion capture methods remain constrained by lighting conditions, limiting their applicability in outdoor scenes.

\subsection{Inertial-based Motion Capture}
Inertial motion capture systems offer a distinct advantage over traditional optical systems by being impervious to occlusions and unrestricted by lighting conditions or recording environment volume.
However, the commercial solutions typically require performers to wear form-fitting suits equipped with a large number of IMUs, leading to setups that are intrusive and cumbersome for the wearer.
Some recent methods~\cite{von2017SIP,huang2018DIP,yi2021transpose,PIPCVPR2022,ren2023lidar} utilize sparse IMUs to produce promising results. Despite it improves the portability of actors during motion capture, it still suffers from drift errors over time and is unable to accurately perceive other physical information, such as human shape and translations.

\subsection{LiDAR-based Motion Capture}
LiDAR technology, known for its precise long-range depth-sensing capabilities, has become increasingly pivotal in fields such as robotics and autonomous driving~\cite{Cong_2022_CVPR,zhu2020ssn,zhu2021cylindrical,Yin2020Centerbased3O,peng2022cl3d}. Its ability to deliver accurate depth information across expansive environments, irrespective of lighting conditions, marks it as an invaluable tool for robust 3D Human Pose and Shape (HPS) estimation. PointHPS~\cite{cai2023pointhps} showcases the potential of using cascaded network architectures for pose and shape estimation directly from point clouds, but the network architecture relies on dense point cloud inputs and is not suitable for sparse point cloud data captured in outdoor large-scale scenes.
LiDARCap~\cite{li2022lidarcap} introduces a graph-based convolutional network approach tailored for interpreting daily human poses within the vast and variable scales of LiDAR-captured scenes. MOVIN~\cite{jang2023movin} explores generative methods for human pose and global translation estimation.
However, these methods focus exclusively on human point clouds in noise-free environments.
Recently, LiveHPS~\cite{ren2024livehps} propose a scene-level human pose and shape estimation by fully utilizing the temporal and spatial information to solve the occlusion and noise disturbance. However, the network tends to take features of all points, including both true human points and noise points, as valid information. The severe noise in data will greatly affect the accuracy of results. Moreover, it only contains the interaction between joints, ignoring the global kinematic information, leading the incoherence in global motion capture outcomes.
Advancing into this research domain, we aim to fully exploit the dynamic and kinematic information available in human movements to capture more coherent and accurate global human motions in noise environments.
\section{Methodology}
\label{sec:method}
\begin{figure*}[t]
	\centering
	\includegraphics[width=\linewidth]{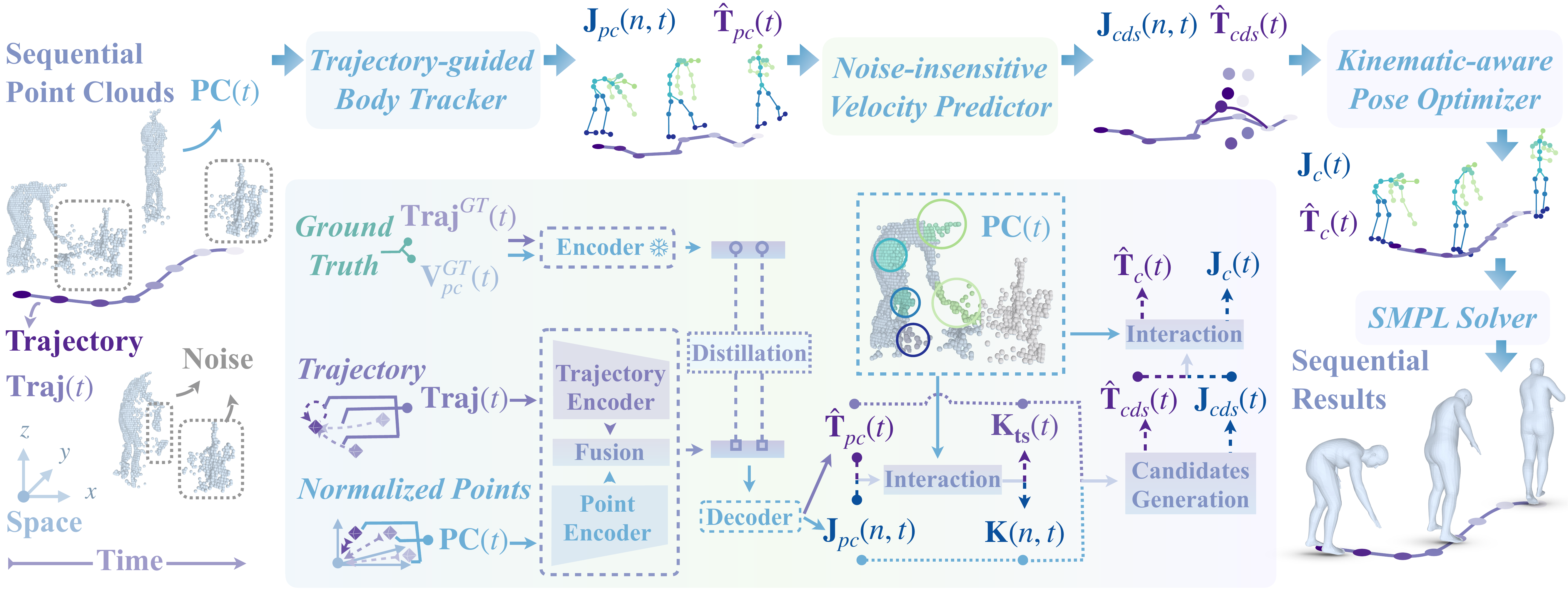}
 \vspace{-2ex}
	\caption{The pipeline of LiveHPS++. It consists of three primary modules, including a trajectory-guided body tracker to predict the human joint and translation, a noise-insensitive velocity predictor to regress the velocity, and the kinematic-aware pose optimizer to enhance the accuracy and coherence of results. Finally, we use SMPL solver to regress the parameters of human poses and shape. Detailed network structure of three modules is also shown under the upper pipeline.}
	\label{fig:pipeline}
	\vspace{-5ex}
\end{figure*}

We introduce LiveHPS++, a novel single-LiDAR-based methodology to estimate the robust and coherent human motions in dynamic free environment. The overview of the pipeline is shown in Fig.~\ref{fig:pipeline}. It takes the sequential noise point clouds as input and aims to acquire the sequential SMPL parameters, including human poses, shapes and translations. The pipeline is structured in three critical components, including trajectory-guided body tracker(Sec~\ref{sec:TBT}), noise-insensitive velocity predictor(Sec~\ref{sec:NVP}), and kinematic-aware pose optimizer(Sec~\ref{sec:KPO}). Firstly, we employ trajectory-guided body tracker to predict the human joints and translations with the assistance of global dynamic information. Secondly, we propose the noise-insensitive velocity predictor to fully utilize the potential association between the human joints and original point cloud for regressing the velocity of each global joint, aiming to optimize the results and concurrently mitigate noise impacts. Then, we design the kinematic-aware pose optimizer to enhance the coherence and accuracy of human motion by predicted velocity. Finally, we use SMPL solver to predict the human poses and shapes from the coherent human joints.

\subsection{Preliminaries}
LiveHPS++ takes sequential single human point clouds as input interspersed with noise from surrounding objects. We resample each input point cloud to a fixed $\mathbf{N}_{input}=256$ by farthest point sample algorithm, then we subtract point clouds with each average location and record the location $\mathbf{Loc}(t)\in R^3$. We define $\boldsymbol{\theta}^{GT}(t) \in \mathbf{R}^{6N_J}$, $\boldsymbol{\beta}^{GT}(t) \in \mathbf{R}^{10}$, and $\mathbf{T}^{GT}(t) \in \mathbf{R}^{3}$ as the ground truth SMPL parameters, $\mathbf{N}_J=24$ and $\mathbf{N}_V=6890$ represents the number of human joint and mesh vertex. We define $\mathbf{PC}(t)$ and $\mathbf{Loc}(t)$ as the normalized point cloud and the mean average positions of raw point cloud, we also follow LIP~\cite{ren2023lidar} to simplify the translation prediction as the offset prediction $\mathbf{\hat{T}}(t)$, and define $\mathbf{\hat{T}}^{GT}(t)$ as the ground truth, the equation is formulated below:
\begin{equation}
    \begin{aligned}
    \mathbf{\hat{T}}^{GT}(t) = \mathbf{Loc}(t)-\mathbf{T}^{GT}(t).
    \end{aligned}
\label{equ:trans2offset}
\end{equation}
We also define the $\mathbf{K}^{GT}(n)$ and $\mathbf{K}_{ts}^{GT}$ as the velocity supervision, the equation is formulated below:
\begin{equation}
    \begin{aligned}
    \mathbf{K}^{GT}(n,t) = &\mathbf{J}^{GT}(n,t+1) - \mathbf{J}^{GT}(n,t),\\
    \mathbf{K}_{ts}^{GT}(t) = &\mathbf{T}^{GT}(t+1) - \mathbf{T}^{GT}(t).
    \end{aligned}
\label{equ:trans2offset}
\end{equation}

\subsection{Trajectory-guided Body Tracker}
\label{sec:TBT}

\begin{figure}[t]
    \centering
      \includegraphics[width=\linewidth]{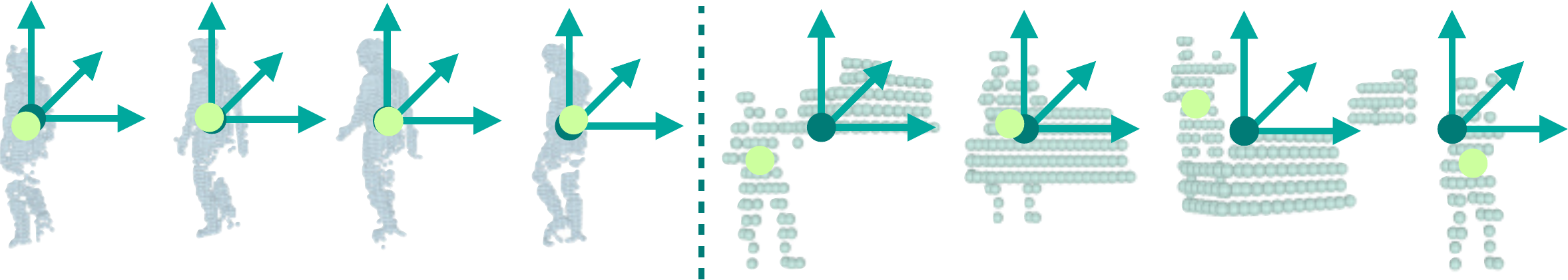}
      \caption{Normalized point cloud. The light green point represents the human root positions, while the dark green point represents the origin of coordinate axis after normalization. The sequential point cloud on the left without noise can be normalized to obtain a relatively stable data distribution, while the data on the right exhibits a more jittery data distribution after normalization due to noise interference.}
    \label{fig:normalized}
    \vspace{-5ex}
\end{figure}

The general strategy of normalization for input data is subtracting the average location, which aims to enhance the generalization capabilities of network and maintain stability in the input data. Notably, the vertex-guided adaptive distillation mechanism, as proposed by LiveHPS~\cite{ren2024livehps}, relies heavily on this normalized input to achieve point representations that closely align with the ground truth vertices, thereby facilitating more accurate and consistent motion capture. However, this conventional normalization strategy encounters significant challenges when dealing with dynamic noise point cloud data. Noise introduced by objects or backgrounds in the scene can lead to substantial fluctuations in the point cloud distribution between adjacent frames, as shown in Fig.~\ref{fig:normalized}. Such fluctuations can disrupt the spatial continuity of the temporal trajectory information, leading to instability in the input data. 
To restore stability in the normalized data amidst noise disruptions, we introduce a dedicated encoder designed to capture trajectory embedding to implicitly model the dynamic characteristics of human movement. We also modify the mechanism as \textbf{vertex-trajectory-guided adaptive distillation} with extra ground truth trajectory information, aiming to fully preserve the trajectory information, thereby enabling to capture of more precise and coherent human motion.
Additionally, the transformation of point cloud features into vertex features in a high-dimensional space potentially predicts the displacement between the point cloud's average position and the true central point. Consequently, we incorporate a decoder branch specifically for predicting translations, further refining the accuracy of our motion capture process.

The distillation mechanism consists of two networks with the same architecture and different input data. We follow LiveHPS to generate the sampling of vertices $\mathbf{V}^{GT}_{pc}(t)$ with consistent distribution with input point cloud and generate the trajectory $\mathbf{Traj}^{GT}(t)$ by Equation~\ref{equ:loc2trj} with $\mathbf{T}^{GT}(t)$. The guidance network takes $\mathbf{V}^{GT}_{pc}(t)$ and $\mathbf{Traj}^{GT}(t)$ as input, we use an MLP encoder to extract the trajectory feature and the PointNet-GRU structure to extract the global point feature. Then we fuse above two features by an MLP layer to get fusion feature $\mathbf{F}_{gt}(t) \in \mathbb{R}^{1024}$ and predict the translations $\mathbf{\hat{T}}_{gt}(t)$ and human joints $\mathbf{J}_{gt}(t)$ by an MLP decoder. We train the guidance network by the mean squared error loss for supervision and freeze the parameters.
\begin{equation}
    \begin{aligned}
    \mathcal{L}_{mse}(\mathbf{J}_{gt})= \sum_t \parallel \mathbf{J}_{gt}(t)-\mathbf{J}^{GT}(t) \parallel_2^2,
    \end{aligned}
\label{equ:loss_VertexJ}
\end{equation}
\begin{equation}
    \begin{aligned}
    \mathcal{L}_{mse}(\mathbf{T}_{gt})= \sum_t \parallel \mathbf{\hat{T}}_{gt}(t)-\mathbf{\hat{T}}^{GT}(t) \parallel_2^2.
    \end{aligned}
\label{equ:loss_VertexT}
\end{equation}
We record the average location $\mathbf{Loc}(t)$ of input data and calculate the trajectory $\mathbf{Traj}(t)$ relative to the first frame input data.
\begin{equation}
    \begin{aligned}
    \mathbf{Traj}(t) = \mathbf{Loc(t)} - \mathbf{Loc(1)}.
    \end{aligned}
\label{equ:loc2trj}
\end{equation}
The learning network takes the input point cloud $\mathbf{PC}(t)$ and trajectory $\mathbf{Traj}(t)$ as input and follows above steps to get the fusion feature $\mathbf{F}_{pc}(t)$ and predict the translations $\mathbf{\hat{T}}_{pc}(t)$ and human joints $\mathbf{J}_{pc}(t)$. The loss function of the trajectory-guided body tracker(TBT) $\mathcal{L}_{TBT}$ is formulated as below:

\begin{equation}
    \begin{aligned}
    \mathcal{L}_{distillation}= \sum_t \mathbf{F}_{gt}(t) \log(\frac{\mathbf{F}_{gt}(t)}{\mathbf{F}_{pc}(t)}),
    \end{aligned}
\label{equ:loss_PC}
\end{equation}

\begin{equation}
    \begin{aligned}
    \mathcal{L}_{TBT}= \lambda_1\mathcal{L}_{distillation}+\lambda_2\mathcal{L}_{mse}(\mathbf{J}_{pc})+\lambda_3\mathcal{L}_{mse}(\mathbf{\hat{T}}_{pc}),
    \end{aligned}
\label{equ:loss_TBT}
\end{equation}
where $\lambda_1=10^3$, $\lambda_2=1$ and $\lambda_3=1$ are hyper-parameters. During inference, the guidance network is not required.
\subsection{Noise-insensitive Velocity Predictor}
\label{sec:NVP}
The trajectory-guided body tracker regresses human joint positions, which leverages skeletal geometric information to enhance motion capture accuracy. This parent-children joint structure allows for the correction of mispredicted joints when partial point cloud data is missing. However, the dependency can lead to error accumulation if the parent joint's prediction is skewed by noisy point cloud data, cause the algorithm noise.
To address the challenge, we aim to enhance the motion feature and learn the kinematic expressions to eliminate the impact of noise. As shown in Fig.~\ref{fig:pipeline}, we design the noise-insensitive velocity predictor to predict the velocity of each human joint which can reflect the kinematic information of human motions and further refine the global pose and translation.

Specifically, the module takes the human joints/translation and input point cloud as the input, utilizes the cross-attention mechanism to make each joint search for truly valuable point features from the original point cloud for feature enhancement, and predicts the velocity $\mathbf{K}(n)/\mathbf{K}_{ts} \in \mathbf{R}^L$ ($L=32$ represents the temporal window size). The loss function is formulated as below:
\begin{equation}
    \begin{aligned}
    \mathcal{L}_{mse}(\mathbf{K}(n))= &\sum_n \parallel \mathbf{K}(n)-\mathbf{K}^{GT}(n) \parallel_2^2,\\
    \mathcal{L}_{mse}(\mathbf{K_{ts}})= &\parallel \mathbf{K}_{ts}-\mathbf{K}_{ts}^{GT} \parallel_2^2.
    \end{aligned}
\label{equ:loss_kt}
\end{equation}
By supervision, the network can learn to distinguish features between real human points and noise points and then eliminate the noise effect.

\subsection{Kinematic-aware Pose Optimizer}
\label{sec:KPO}
Leveraging the velocity values derived previously, we develop a kinematic-aware optimizer to further correct motion outcomes. 
The predicted velocity provides the temporal connection of each joint between $t+1$ frame and $t$ frame, while the predicted joints $\mathbf{J}_{pc}(n,t)$ provide joint-wise spatial position of each frame, thus we can generate candidate joints $\mathbf{J}_{cds}(n,t,t)$ by:
\begin{equation}
    \begin{aligned}
    \mathbf{J}_{cds}(n_i,t_i,t_j) = \mathbf{J}_{pc}(n_i,t_i) + \Delta t\sum_{t={t_i}}^{t_j} \mathbf{K}(n_i,t),\\
    \mathbf{\hat{T}}_{cds}(t_i,t_j) = \mathbf{Loc}(t) - (\mathbf{T}_{pc}(n_i,t_i) + \Delta t\sum_{t={t_i}}^{t_j} \mathbf{K}_{ts}(t)).
    \end{aligned}
\label{equ:gen_cds}
\end{equation}
$\mathbf{J}_{cds}(n_i,t_i,t_j)$ represents the $t_i$ candidate joints for $n_i$ joint in $t_j$ frame, thereby we can generate $(L-1)$ candidate joints. Considering that long-term kinematic optimization causes inaccurate results at the extreme points of one motion sequence and causes cumulative errors over time, meanwhile, short-term kinematic optimization with short-change information can only refine minority mutation cases and can not keep the coherence of the whole sequence. We design the module with cross-attention architecture to build the connection between the candidate joints with the original input data, facilitating the extraction of more pronounced features and further benefiting more accurate joint correction. Finally, we get the coherent and accurate global human joints $\mathbf{J}_{c}(t)$ and translations $\mathbf{T}_{c}(t)$, the loss function of the kinematic-aware pose optimizer(KPO) $\mathcal{L}_{KPO}$ is formulated as below:
\begin{equation}
    \begin{aligned}
    \mathcal{L}_{KPO} = \lambda_4\mathcal{L}_{mse}(\mathbf{J}_{c})+\lambda_5\mathcal{L}_{mse}(\mathbf{\hat{T}}_{c}),
    \end{aligned}
\label{equ:loss_pko}
\end{equation}
where $\lambda_4=1$ and $\lambda_5=1$ are hyper-parameters.
\subsection{SMPL Solver}
\label{sec:AFS}
In the last stage, we follow the LiveHPS~\cite{ren2024livehps} to use the attention-based SMPL solver to predict the human poses $\theta(t)$ and shape $\beta$. Finally, we use SMPL model to generate the human joints and mesh vertex as below:
\begin{equation}
    \begin{aligned}
    \hat{\mathbf{J}}_{smpl}(t),\hat{\mathbf{V}}_{smpl}(t) = \operatorname{SMPL}(\theta(t),\beta,\mathbf{\hat{T}}_{c}(t)).
    \end{aligned}
\label{equ:smpl}
\end{equation}
The loss function of the SMPL solver $\mathcal{L}_{smpl}$ is formulated as:
\begin{equation}
    \begin{aligned}
    \mathcal{L}_{smpl} = &\lambda_6\mathcal{L}_{mse}(\mathbf{J}_{smpl})+\lambda_7\mathcal{L}_{mse}(\mathbf{V}_{smpl})\\+&\lambda_8\mathcal{L}_{mse}(\theta)+\lambda_{9}\mathcal{L}_{mse}(\beta),
    \end{aligned}
\label{equ:solver_loss}
\end{equation}
where $\lambda_6=\frac{100}{N_j}$, $\lambda_7=\frac{100}{N_v}$, $\lambda_8=1/5$ and $\lambda_{9}=1$ are hyper-parameters. It is worth noting that due to the large number of noise points in the input data, the SUCD loss proposed by LiveHPS is not suitable.
\section{Dataset}
\label{sec:dataset}
\begin{figure*}[t]
	\centering
	\includegraphics[width=\linewidth]{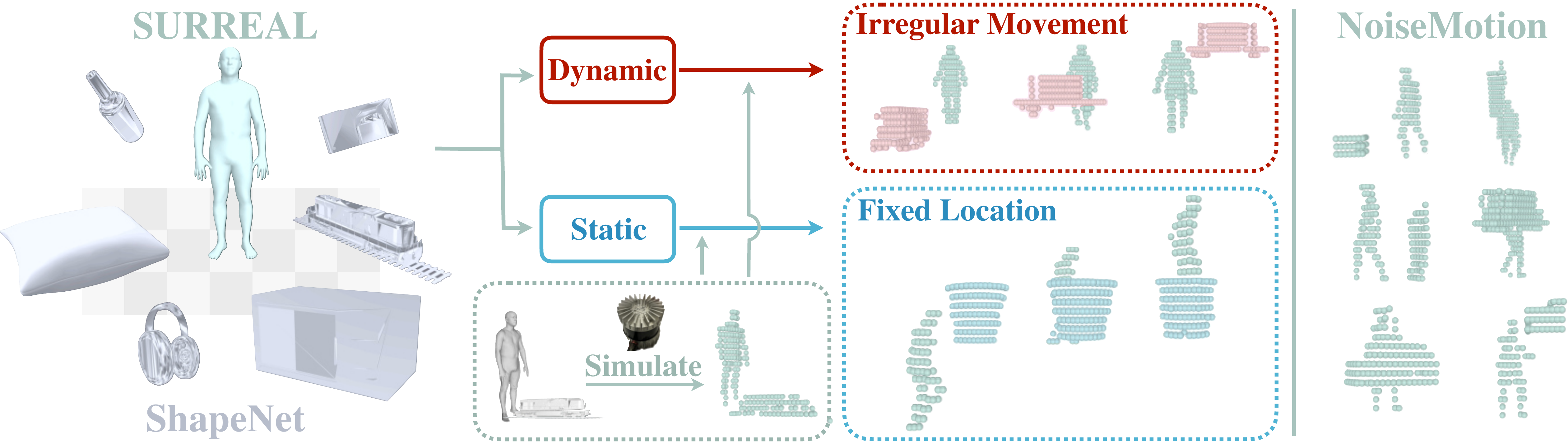}
 \vspace{-2ex}
	\caption{The NoiseMotion dataset simulation pipeline, integrating dynamic human motion and static object noise to simulate real-world human-object interactions.}
	\label{fig:noisemotion}
	\vspace{-5ex}
\end{figure*}
Previous LiDAR-based methods use extensive synthetic data in training to enhance the generalization of network, however, existing synthetic data primarily simulate arbitrary sensing noise and exhibit random human point cloud translations across sequences, failing to accurately reflect the complexities encountered in real-world scenarios.
Acknowledging the limitations of existing synthetic datasets, we propose the NoiseMotion, which leverages human motion data from SURREAL~\cite{varol2017learning} and 3D object models from ShapeNet~\cite{chang2015shapenet} to meticulously simulate complex noise patterns resulting from human-object interactions.
The pipeline of data augmentation is shown in Fig.~\ref{fig:noisemotion}, which categorizes noisy objects as either dynamic or static. Dynamic objects, which can appear unexpectedly and significantly alter the point cloud distribution, contrast with static objects that introduce noise through human-object interaction or proximity, a prevalent noise type in everyday settings. To further diversify the noise types and their distribution, we employ data augmentation techniques like random rotation and scaling. This approach is critical for accurately reflecting real-world complexities, significantly boosting the network's generalization capability.
Compared to the real human motion dataset FreeMotion-OBJ~\cite{ren2024livehps}, NoiseMotion offers a vastly richer collection, including 51,300 unique 3D object models from ShapeNet and 1,021,802 human motions from SURREAL. In contrast, FreeMotion-OBJ provides fewer than ten types of dynamic objects. This stark difference underscores NoiseMotion's importance and necessity for advancing LiDAR-based applications.
\section{Experiment}
\label{sec:experiment}
\begin{figure}[ht!]
	\centering
	\includegraphics[width=\linewidth]{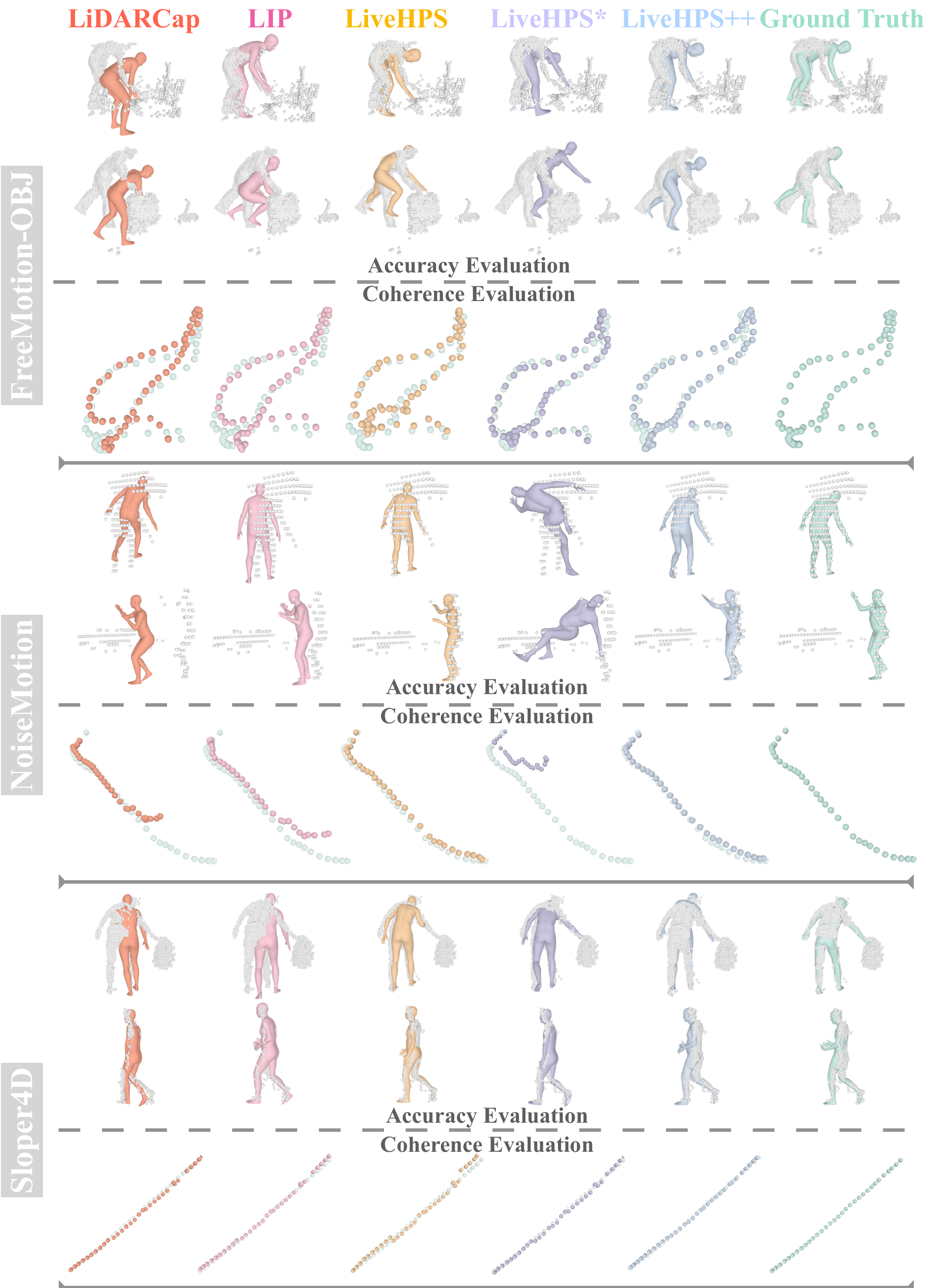}
        \vspace{-4ex}
	\caption{Qualitative comparisons. The point cloud matches the result better, representing more accurate estimation for pose, shape, and translation. Each point in the visualization of coherence evaluation represents the frame-wise global human translations in the bird's-eye view.}
	\label{fig:compare}
	\vspace{-5ex}
\end{figure}

In this section, we present comprehensive experiments to validate the effectiveness, robustness, and coherence of our method, LiveHPS++, against current state-of-the-art (SOTA) methods, including LiDARCap~\cite{li2022lidarcap}, LIP~\cite{ren2023lidar}, and LiveHPS~\cite{ren2024livehps}. Additionally, we present detailed ablation studies to assess the contribution of our network architecture's components. Following LiveHPS, our evaluation metrics include J/V Err(PS/PST)($mm$) and Ang Err($degree$). Notably, scene-level unidirectional Chamfer distance in millimeters(SUCD) is not suitable for noisy input data, so we don't use SUCD for metrics.
To provide deeper insight into the effect of our method on the accuracy and coherence of human motion trajectories, we introduce two additional metrics: 1) Acceleration Error(Accel Err)($m/s^2$)$\downarrow$: which quantifies the mean acceleration error across global human joints, calculated against ground truth data to gauge the trajectory accuracy of human motion; 2) Jitter($10^2m/s^3$)$\downarrow$: which evaluates the average jerk across global human joints, assessing coherence of motion trajectories independently of ground truth data and offering a measure of the fluidity of captured movements.


\subsection{Implemantation Details}
Our network structure is implemented by PyTorch version 1.10.0 and CUDA 11.4, trained over 200 epochs with batch size of 64, sequence length of 32, and learning rate of $10^{-3}$, on an Intel(R) Xeon(R) Gold 5318Y CPU and 4 NVIDIA A40 GPUs. Other training configuration aligns with the settings established by LiveHPS. As for the dataset, we use FreeMotion~\cite{ren2024livehps}, Sloper4D~\cite{dai2023sloper4d}, and our NoiseMotion. FreeMotion and Sloper4D are LiDAR-based human motion datasets, some of the data contains noise points from objects. Our NoiseMotion dataset is based on the SURREAL~\cite{varol2017learning} and ShapeNet~\cite{chang2015shapenet}, we generate the synthetic data consisting of human motion dataset SURREAL and object dataset ShapeNet, which simulate the challenge human-object interaction case in dynamic free environment with severe noise. The dataset splitting is followed by LiveHPS. All methods are trained on the training set of NoiseMotion, FreeMotion, and Slopher4D.
\subsection{Comparison}
\begin{table}[t!]\small
\centering
\caption{Comparison with state-of-the-art methods on various datasets. Lower values represent better performance for all metrics. FreeMotion-OBJ means the human-object interaction part of FreeMotion. Notably, LiveHPS$^*$ is trained on the real dataset and previous clean synthetic data, which contains the same human motion as the NoiseMotion.}
\vspace{-2ex}
\resizebox{\linewidth}{!}{
\setlength\tabcolsep{3pt} 
\begin{tabular}{c|ccccc|ccccc}
\toprule
\multirow{2}{*}{} & \multicolumn{5}{c|}{NoiseMotion~\cite{varol2017learning}~\cite{chang2015shapenet}}& \multicolumn{5}{c}{FreeMotion-OBJ~\cite{ren2024livehps}}\\
\cmidrule(r){2-6}
\cmidrule(r){7-11}
& J/V Err(PS)$\downarrow$& J/V Err(PST)$\downarrow$&Ang Err$\downarrow$&Accel Err$\downarrow$&Jitter$\downarrow$
& J/V Err(PS)$\downarrow$& J/V Err(PST)$\downarrow$&Ang Err$\downarrow$&Accel Err$\downarrow$&Jitter$\downarrow$\\
\midrule
LiDARCap~\cite{li2022lidarcap}&52.63/64.65&400.66/402.58&10.87&42.48&765.89
&84.11/100.61&181.82/189.32&16.61&7.21&62.47\\
LIP~\cite{ren2023lidar}&62.41/77.97&192.79/198.66&14.07&25.31&451.74
&87.50/108.28&158.38/170.90&20.16&7.09&60.19\\
LiveHPS~\cite{ren2024livehps}&48.37/60.42&74.70/83.84&12.19&5.78&68.65
&70.73/88.43&146.78/158.00&17.81&8.82&117.79\\
LiveHPS$^*$~\cite{ren2024livehps}&370.29/432.93&561.49/611.40&27.32&49.74&884.24
&83.33/101.70&133.82/146.12&16.84&8.38&100.82\\
\midrule
\textbf{Ours}&\textbf{34.00/42.75}&\textbf{58.53/64.51}&\textbf{10.63}&\textbf{3.48}&\textbf{59.35}
&\textbf{58.11/72.55}&\textbf{128.60/136.94}&\textbf{15.85}&\textbf{7.01}&\textbf{30.96}\\
\midrule
\midrule
\multirow{2}{*}{} & \multicolumn{5}{c|}{FreeMotion~\cite{ren2024livehps}}& \multicolumn{5}{c}{Sloper4D~\cite{dai2023sloper4d}}\\
\cmidrule(r){2-6}
\cmidrule(r){7-11}
& J/V Err(PS)$\downarrow$& J/V Err(PST)$\downarrow$&Ang Err$\downarrow$&Accel Err$\downarrow$&Jitter$\downarrow$
& J/V Err(PS)$\downarrow$& J/V Err(PST)$\downarrow$&Ang Err$\downarrow$&Accel Err$\downarrow$&Jitter$\downarrow$\\
\midrule
LiDARCap~\cite{li2022lidarcap}&86.28/104.17&180.36/188.58&15.51&6.28&70.57
&71.64/84.23&138.71/147.79&13.72&6.16&88.50\\
LIP~\cite{ren2023lidar}&85.49/104.05&141.36/153.25&19.73&6.16&68.11
&74.38/91.89&134.69/146.90&20.53&6.59&96.16\\
LiveHPS&74.71/90.79&130.41/141.08&16.96&7.27&85.38
&53.37/63.15&88.35/95.85&13.08&5.88&73.56\\
LiveHPS$^*$&69.38/83.86&119.22/128.55&15.80&6.99&86.07
&48.28/59.02&77.73/85.83&12.77&5.64&97.41\\
\midrule
\textbf{Ours}&\textbf{61.91/75.27}&\textbf{112.13/120.39}&\textbf{15.40}&\textbf{5.42}&\textbf{33.16}
&\textbf{42.70/50.62}&\textbf{76.98/81.67}&\textbf{11.92}&\textbf{4.34}&\textbf{59.97}\\
\bottomrule
\end{tabular}}
\label{tab:compare}
\vspace{-5ex}
\end{table}
We evaluate LiveHPS++ on the testing sets of NoiseMotion, FreeMotion, and Sloper4D. To specifically assess our network's resilience to noise, we additionally show the evaluation results on the human-object interaction sequences from the FreeMotion testing set, referred to as FreeMotion-OBJ.
This allows for a focused assessment of noise-handling capabilities.
LiveHPS++ is benchmarked against leading LiDAR-based methods to underscore its state-of-the-art (SOTA) performance, as detailed in Tab.~\ref{tab:compare}.
The experimental results demonstrate our LiveHPS++'s exceptional performance across various metrics, with notable advancements in Acceleration Error (Accel Err) and Jitter especially in NoiseMotion, underscoring our method's adeptness at handling dynamic motion coherence in complex environments with severe noise.
We can observe that, after training on NoiseMotion, LiveHPS exhibits enhanced performance on NoiseMotion and FreeMotion-OBJ dataset characterized by high noise levels compared with LiveHPS$^*$, which is trained on real data and previous clean synthetic dataset, which provides the same human motion as the NoiseMotion. This demonstrates the value of our synthetic data NoiseMotion.
However, LiveHPS can not achieve stable generalizability and shows a relative decline in FreeMotion and Sloper4D when compared to LiveHPS$^*$.
Our LiveHPS++ can achieve SOTA performance in both cases with severe noise or not, highlighting the robustness and generalization capabilities of our method.


The qualitative comparisons in Fig.~\ref{fig:compare} further accentuate LiveHPS++'s proficiency in sustaining stability and delivering coherent outcomes under significant noise conditions. Through visualizations of global human motion by each method on the test datasets, LiveHPS++'s superior noise immunity is evident. For instance, in challenging scenarios such as those presented in FreeMotion-OBJ and NoiseMotion, our LiveHPS++ reliably differentiates between human-related and noise points, unlike LiveHPS$^*$, which misinterprets noise as legitimate motion cues in noise environments, leading to inaccuracies. While other methods trained on NoiseMotion dataset show some capacity to disregard irrelevant noise, their performance is still noticeably impacted. Moreover, in scenarios with occlusion, such as those within the Sloper4D's second row where the hand is occluded, LiveHPS++ consistently outperforms competing methods in these scenarios. 
As Fig.~\ref{fig:teaser} shows, thanks to our effective network design which can implicitly and explicitly model dynamic and kinematic features of human motions, our LiveHPS++ can capture coherent and accurate human motion, even in a real-time captured noise scenario.

We also visualize the bird-eye-view of global human translations in Fig.~\ref{fig:compare}. Our method showcases superior performance in predicting both accurate and coherent global human translations, a capability that is particularly evident in handling the complex and challenging motions in FreeMotion-OBJ dataset. In contrast, LiDARCap derives translation estimates from the average locations within point clouds, which can result in inaccuracies and jerkiness due to variations in point distribution, occlusions, and noise interference. Although LiveHPS effectively leverages temporal and spatial data for enhanced accuracy, it falls short of maintaining coherence across its predictions. Our method, on the other hand, achieves both coherent and precise global translations.

\subsection{Ablation Study}
\begin{table}[t]
\centering
\caption{Ablation studies for our network modules on FreeMotion-OBJ. We also evaluate the internal details of each module.}
\vspace{-2ex}
\resizebox{\linewidth}{!}{
\setlength\tabcolsep{5pt} 
\begin{tabular}{c|c|ccccc}
    \toprule
    \multicolumn{2}{c|}{} &J/V Err(PS)$\downarrow$& J/V Err(PST)$\downarrow$&Ang Err$\downarrow$&Accel Err$\downarrow$&Jitter$\downarrow$\\ 
    \cmidrule(r){1-2}
    \cmidrule(r){3-7}
    \multirow{2}{*}{Network Module}& w/o TBT&71.68/90.08&153.65/164.79&17.52&7.29&33.89\\
    & w/o NVP\&KPO & 68.37/84.92&134.11/145.01&17.23&7.79&71.82\\
    \midrule
    \multirow{2}{*}{Trajectory-guided Body Tracker}&frame-wise&68.57/86.15&151.73/162.15&16.69&7.54&43.53\\
    &sequence-wise&85.42/106.91&165.71/179.41&19.40&7.10&31.32\\
    \midrule
    \multirow{2}{*}{Kinematic-aware Pose Optimizer}& short-term optimizer&64.04/79.27&132.38/141.57&16.15&7.63&55.50\\
    &long-term optimizer&68.06/83.62&150.90/159.71&16.71&7.21&42.51\\
    \midrule
    \multirow{3}{*}{Translation Estimation}& Average&-&177.13/183.48&-&8.18&94.73\\
    &LIP&-&132.82/140.94&-&8.85&93.56\\
    &LiveHPS&-&130.56/138.71&-&8.12&78.09\\
    \midrule
    \multicolumn{2}{c|}{Ours}&\textbf{58.11/72.55}&\textbf{128.60/136.94}&\textbf{15.85}&\textbf{7.01}&\textbf{30.96}\\
    \bottomrule
\end{tabular}}
\vspace{-4ex}
\label{tab:ablation-s}
\end{table}

To prove the superiority of each network module in our LiveHPS++, we conduct ablation study for the network architecture on FreeMotion-OBJ to demonstrate the effectiveness of each module and we also evaluate more details for each module as shown in Tab.~\ref{tab:ablation-s}.

\noindent\textbf{Network Architecture.} The network without trajectory-guided body tracker(TBT) yields coherent yet inaccurate results compared to the network without noise-insensitive velocity predictor(NVP) and kinematic-aware pose optimizer(KPO). This discrepancy highlights the critical role of the TBT module in enhancing the accuracy of local motion within noisy environments and the effectiveness of KPO module in optimizing the coherence of sequential motions and translations. Together, the results underscore the significant improvements in both accuracy and coherence brought about by the integration of TBT and KPO in our LiveHPS++'s network architecture.

\noindent\textbf{Trajectory-guided Body Tracker.} We conduct the detailed ablation study on Trajectory-guided Body Tracker(TBT) module by exploring the performance of frame-wise and sequence-wise data normalization. Frame-wise normalization, which does normalization by subtracting point clouds with the average location of each frame, is often utilized in previous LiDAR-based human motion caption methods. It is beneficial for mitigating sensitivity to scale and position variances and network convergence acceleration, achieving accurate prediction but but falls short by omitting essential physical movement information. Sequence-wise normalization, which is sequential normalization by subtracting point clouds with the average location of the first frame, can retain the real-world physical trajectory, achieving smooth but inaccurate results. Our trajectory-guided body tracker facilitates both reduced sensitivity to scale and positional differences, while retaining real-world physical motion information, thus allowing accurate and global human motion results to be obtained.


\noindent\textbf{Kinematic-aware Pose Optimizer.} 
The Kinematic-aware Pose Optimizer (KPO) module refines human joint and translation predictions using velocities predicted by a noise-insensitive velocity predictor. It integrates both short-term and long-term kinematic information for joint-wise optimization. We contrast our method with two others: a short-term approach, which optimizes each frame with the result and the velocity of the previous frame, and a long-term strategy, which optimizes the entire sequence using the results of the first frame and velocity. The short-term optimizer enhances adjacent frame coherence but introduces jerkiness in long sequences and overlooks long-range coherence. Conversely, the long-term optimizer maintains overall coherence but leads to accumulated errors and dependency on the initial frame's accuracy. Our KPO can achieve accurate and coherent results by considering both short-term and long-term kinematic optimization.

\noindent\textbf{Translation Estimation.}
We further refine human translation estimation, achieving superior accuracy and coherence by minimizing noise impact through trajectory embedding in the TBT module and enhancing translation prediction in the KPO module by leveraging temporal dynamics. This approach significantly surpasses translation estimation methods proposed in LIP and LiveHPS in both acceleration error and jitter metrics, showcasing our method's advanced capability in capturing precise and fluid human motion translations.

\begin{figure}[ht!]
 \vspace{-4ex}
	\centering
	\includegraphics[width=\linewidth]{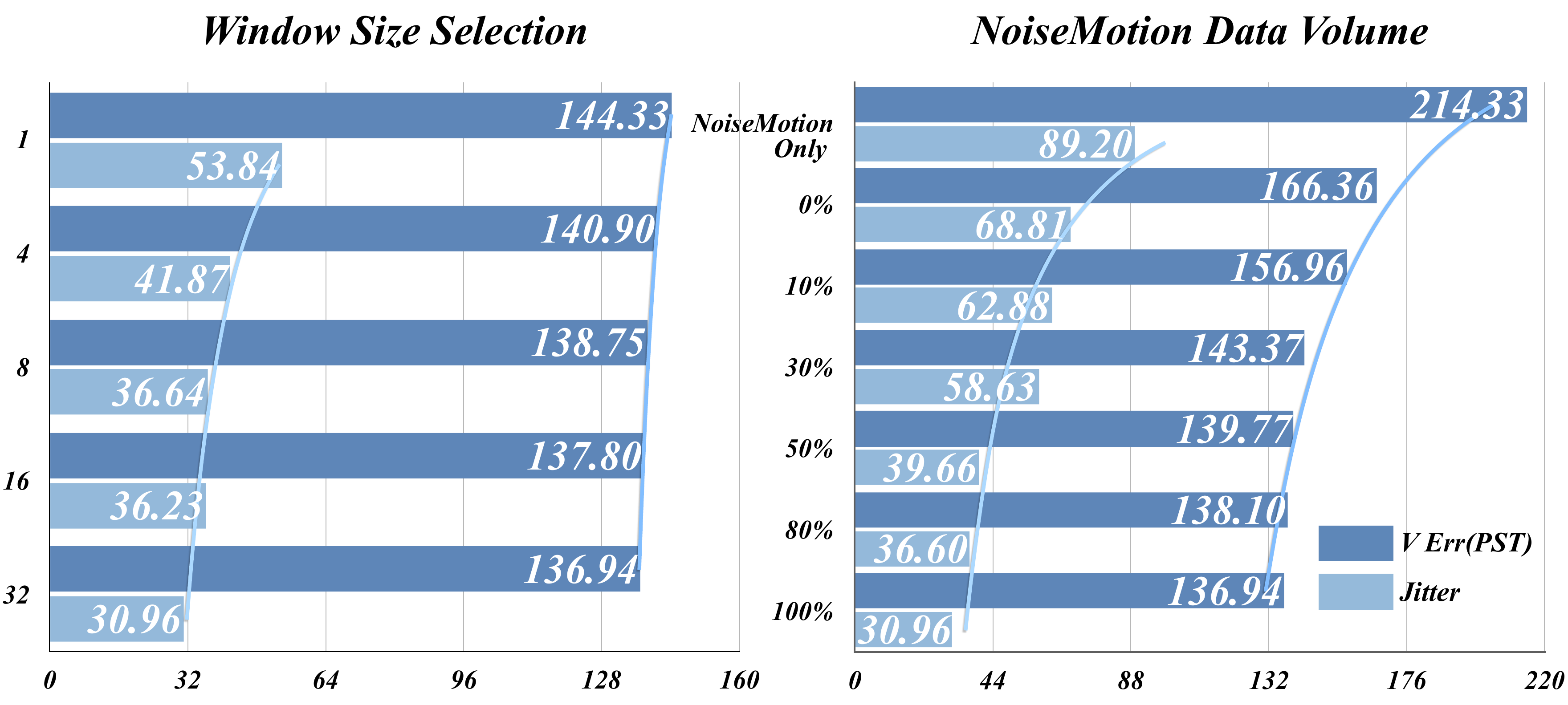}
        \vspace{-4ex}
	\caption{Ablation study for the temporal window size selection and evaluate the impact of NoiseMotion volume leveraged for training on model performance.}
	\label{fig:ablation}
	\vspace{-5ex}
\end{figure}
\noindent\textbf{Window Size Selection of KPO.}
The selection of an optimal temporal window size within the Kinematic-aware Pose Optimizer (KPO) module is crucial for enhancing motion and translation coherence. We experiment with various window sizes 1, 4, 8, 16, and 32. Observing that a window size of 32 yields the best results, as depicted in Figure~\ref{fig:ablation}. This configuration led to a slight but consistent decrease in Vertex Error (V Err) and a more pronounced reduction in Jitter, indicating more accurate and coherent sequences. This highlights the significance of a suitable receptive field for optimal sequence coherence and accuracy and confirms our algorithm's adaptability to various temporal lengths.

\noindent\textbf{NoiseMotion Data Volume.}
We use NoiseMotion to enhance the generalization ability of the network, especially when dealing with noisy data. We discuss the impact of the volume of the synthetic data on network performance in Fig.~\ref{fig:ablation}. When we only use NoiseMotion for training, there still exists a domain gap between the real data and synthetic data. We gradually add the NoiseMotion data volume(0\%, 10\%, 30\%, 50\%, 80\%, and 100\%) to real data for training, and the performance gradually improves, which demonstrates the synthetic data is significant for the task.
\label{sec:discussion}
\section{Conclusion}

In this paper, we introduce a novel and effective single-LiDAR-based approach, distinguished by its ability to precisely capture accurate and coherent 3D human motions across various unconstrained environments. By fully harnessing the dynamic and kinematic attributes derived from global human movements, we effectively mitigate the adverse effects of significant noise. Additionally, we present a new synthesized motion dataset aimed at augmenting the network's adaptability in noisy conditions. Comprehensive experiments demonstrate the obvious superiority of our method, particularly in terms of local pose accuracy and global pose coherence, rendering our technique highly suitable for practical applications.


%
%
\bibliographystyle{splncs04}
\bibliography{main}
\end{document}